# PQTNet: Pixel-wise Quantitative Thermography Neural Network for Estimating Defect Depth in Polylactic Acid Parts by Additive Manufacturing

Lei Deng [a], Wenhao Huang [a], Chao Yang [b], Haoyuan Zheng [b], Yibin Tian [a], Yue Ma (✉)[b]

[a] College of Mechatronics and Control Engineering & State Key Lab of Radio Frequency Heterogenous Integration, Shenzhen University, Shenzhen 518060, China

[b] School of Robotics, Xi'an Jiaotong-Liverpool University, Suzhou 215123, China

## ABSTRACT

Defect depth quantification in additively manufactured (AM) components remains a significant challenge for non-destructive testing (NDT). This study proposes a Pixel-wise Quantitative Thermography Neural Network (PQT-Net) to address this challenge for polylactic acid (PLA) parts. A key innovation is a novel data augmentation strategy that reconstructs thermal sequence data into two-dimensional stripe images, preserving the complete temporal evolution of heat diffusion for each pixel. The PQT-Net architecture incorporates a pre-trained EfficientNetV2-S backbone and a custom Residual Regression Head (RRH) with learnable parameters to refine outputs. Comparative experiments demonstrate the superiority of PQT-Net over other deep learning models, achieving a minimum Mean Absolute Error (MAE) of 0.0094 mm and a coefficient of determination ($R^2$) exceeding 99%. The high precision of PQT-Net underscores its potential for robust quantitative defect characterization in AM.

## 1. INTRODUCTION

Additive Manufacturing (AM) holds great potential in aerospace, biomedical, and automotive manufacturing [1-3]. AM technology constructs solid components layer by layer based on three-dimensional Computer-Aided Design (CAD) models through slicing algorithms [4], utilizing techniques such as Fused Deposition Modeling (FDM), St[1]ereolithography (SLA), and Selective Laser Sintering (SLS). A wide range of materials, such as Polylactic Acid (PLA), Acrylonitrile Butadiene Styrene (ABS), Polyethylene Terephthalate Glycol (PETG), Nylon, and Carbon Fiber Reinforced Polymer (CFRP) [5, 6] have been utilized in the AM technology to realize various functions. PLA, in particular, has emerged as a research hotspot in AM due to its excellent printability, favorable mechanical properties, and biodegradable nature, aligning well with the principles of green manufacturing and sustainable development [7-9].

However, PLA AM faces several technical challenges due to non-equilibrium thermal cycling during processing [10], Key defects include: (1) porosity from insufficient interlayer bonding, significantly affecting mechanical properties and structural integrity [11], (2) microcracks originating from thermal stresses and interlayer shrinkage mismatches, creating fatigue failure initiation sites; (3) heterogeneous inclusions from impurities, molecular degradation, or contamination, disrupting matrix continuity; and (4) non-uniform crystallinity distribution from unstable thermal fields and variable cooling rates. The synergistic effects of porosity and crystallinity critically determine final component performance [12], often resulting in material anisotropy and dimensional instability. These internal flaws substantially compromise mechanical performance and durability, potentially triggering failure mechanisms that threaten product reliability and safety [13].

Therefore, developing efficient techniques for the detection of internal defects in PLA components by AM, along with accurate assessment of defect depth, is of significant scientific and engineering importance. Such advancements are essential for improving the quality control of AM-fabricated PLA components and expanding their applications. Non-Destructive Testing (NDT) technologies [14-16] play an important role in quality assessment of AM-fabricated parts. Among them, active infrared thermography [17-19] has emerged as a highly promising non-contact NDT technique, characterized by rapid detection, large-area coverage, and the capability for quantitative analysis. These advantages make it an effective technological solution for defect detection and evaluation in AM.

Classical defect depth estimation methods utilize physical modeling and feature parameter extraction. One category exploits correlations between characteristic thermal response times (peak time, slope time, derivatives) and defect depth. For example, Wei et al. [20] proposed an improved peak slope time method by integrating pulse

---

**CONTACT:** Yue Ma: Yue.Ma02@xjtlu.edi.cn; Chao Yang: Chao.Yang.xjtlu@gmail.com; Haoyuan Zheng: Haoyuan.Zheng.xjtlu@gmail.com;

thermography with CFRP composite specimens. By establishing a correlation model among defect size, sample thickness, and characteristic time. Additionally, Wei et al. [21] investigated several methods based on Specific Characteristic Times (SCT), including Peak Contrast Time (PCT), Peak Slope Time (PST), Logarithmic Second Derivative (LSD), and Absolute Peak Slope Time (APST).

The rapid advancement of ML and Deep Neural Networks (DNNs) has introduced novel solutions in thermography NDT [22]. The integration of these approaches offers innovative pathways toward intelligent detection and evaluation. Dong et al. [23] introduced a spatiotemporal 3-D Convolutional Neural Network (CNN). By integrating spatial-temporal convolutional filters with Group Normalization (GN) techniques, it enabled simultaneous detection and depth estimation of subsurface defects in CFRP materials. Wei et al. [24] constructed an open-access pulsed thermography dataset and adopted the YOLOv5 DNN to achieve simultaneous prediction of defect location and depth, providing an effective avenue for the development of data-driven intelligent NDT. Multimodal and feature-level fusion approaches have been increasingly explored in thermography to enhance defect detection by leveraging complementary spatial or temporal-based information. For example, Mohammed et al. [25] proposed a multimodal attention network for pulsed thermography named PT-Fusion, which integrates features from Principal Component Analysis (PCA) and Thermographic Signal Reconstruction (TSR). This approach enabled subsurface defect segmentation and depth estimation.

Based on the above analysis, several key challenges remain in the detection and quantitative analysis of internal defects in AM-fabricated components: (1) The existing defect depth estimating models lack improvements specifically targeting the temporal characteristics of infrared signals, though it has been utilized for subsurface defect detection in active thermography [26-29]; (2) the proposed models often exhibit limited capacity for continuous value prediction of subsurface properties; and (3) The absence of large-scale annotated datasets in active thermography poses a significant limitation on the model's learning capacity.

This study aims to design a network architecture tailored for active thermography to achieve continuous and high-precision prediction of subsurface defects in AM-fabricated components. Specifically, it addresses the following key technical challenges: first, to construct a DNN architecture capable of effectively processing temporal infrared thermography data; second, to develop a defect prediction algorithm adapted to the measurement of the continuously changed depth of internal defects.

## 2. METHODS

### 2.1. Specimen

PLA is used for defect depth prediction in this study, because it possesses favorable mechanical properties and processability, while being fully biodegradable into non-toxic substances. Thermophysical characteristics of the PLA, such as the thermal conductivity, specific heat capacity, and emissivity, along with several commonly used materials, are summarized in Table 1. PLA exhibits relatively low thermal conductivity, which results in slower heat transfer. This property facilitates the formation of distinct temperature gradients between defective and non-defective regions, thereby creating favorable conditions for thermal imaging-based defect detection.

**Table 1** Characteristic data of some plastic materials.

| Material | Thermal Conductivity (W/m·K) | Specific Heat (kJ/kg·K) | Density (g/cm³) | Emissivity |
|---|---|---|---|---|
| ABS | 0.18 - 0.25 | 1.3 - 1.8 | 1.04 | 0.90- 0.95 |
| PA | 0.25 - 0.30 | 1.7 - 2.0 | 1.10-1.15 | 0.90- 0.95 |
| PC | 0.19 - 0.22 | 1.2 - 1.4 | 1.20 | 0.90 |
| PLA | 0.13 - 0.25 | 1.8 - 2.0 | 1.20-1.25 | 0.90- 0.95 |

To fabricate samples with varying defect depths, a CREALITY 3D printer (Model: K2 PLUS) was employed. This industrial-grade equipment provides a printing accuracy of up to 0.1 mm and supports a maximum build volume of 350 mm × 350 mm × 350 mm. The PLA specimens were printed with dimensions of 90 mm × 90 mm × 5 mm. Circular defects with a fixed radius of 8 mm were embedded, with defect depths ranging from 0.24 mm to 1.52 mm. Fig.1 presents schematic diagrams of the front and side views of the specimens, along with the distribution of defect depths.

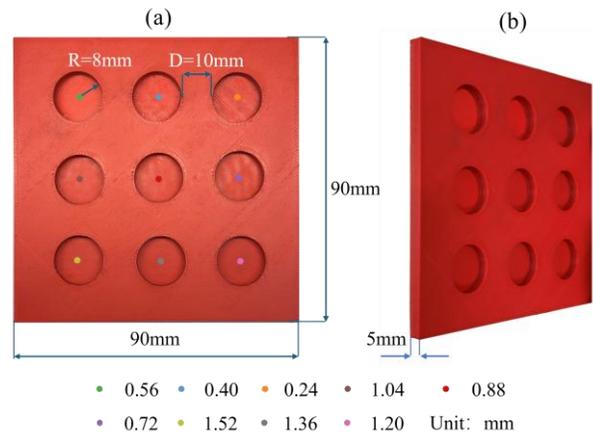

**Fig. 1** The front view (a) and side view (b) of the AM-fabricated PLA parts along with the defect depths indicated by the colored dots in the middle of the circular defects.

### 2.2. Experiment setup

As shown in Fig.2, the experimental setup comprised two 800W halogen lamps with mechanical timer control, an IR camera, and display monitor. The MV-CI003-GL-N6 IR camera (Hikvision) featured 640×480 pixel resolution, 35 mK thermal sensitivity, and 88.5°×73.2° FOV. Video acquisition used Hikvision MVS software. The camera was positioned vertically 30 cm from the sample center, with lamps placed symmetrically at 45° angles for uniform illu-

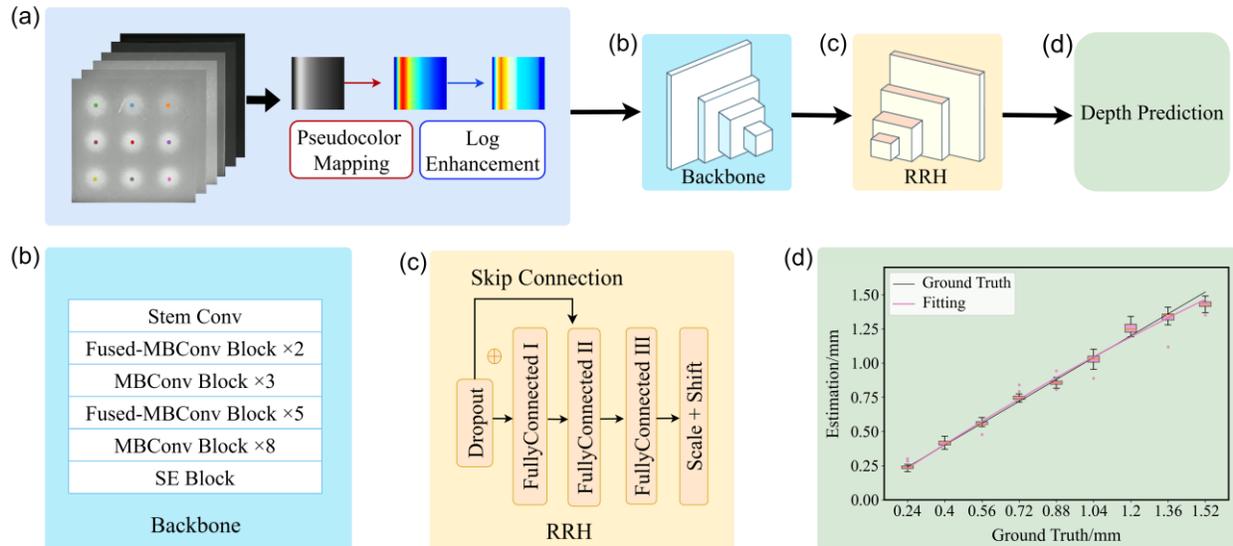

**Fig. 3** (a) the architecture of the PQTNet. (b) EfficientNetV2-S backbone with various convolutional blocks and SE attention; (c) a customized Residual Regression Head (RRH) module. (d) Example of depth prediction results.

mination and minimal reflection artifacts. Samples underwent 30-second long-pulse thermal excitation. The IR camera recorded at 50 Hz for 220 seconds, capturing 11,000 infrared temporal frames.

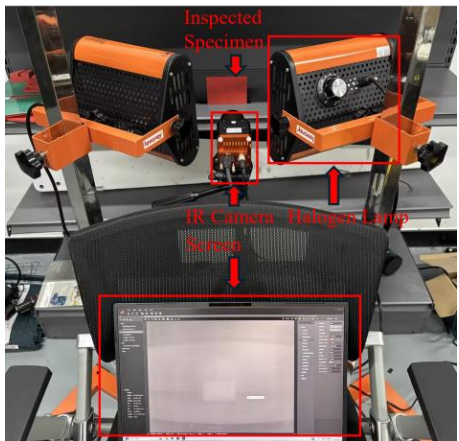

**Fig. 2** Overall experimental setup.

### 2.3. The deep neural network (PQTNet)

*2.3.1. The overall network architecture*

Fig.3 illustrates the overall architecture of the proposed Pixel-wise Quantitative Thermography Neural Network (PQTNet) for defect depth estimation in this study. It consists of three main components: a backbone encoder based on the pretrained EfficientNetV2-S [30], a channel attention mechanism module (SE Block), and a customized Residual Regression Head (RRH).

EfficientNetV2-S was selected as the backbone encoder due to the following advantages: (1) Its progressive learning strategy and adaptive regularization enhance generalization performance, especially on small datasets; (2) Compared to traditional CNN architectures, EfficientNetV2-S excels at capturing fine-grained features, making it particularly suitable for thermal image regression tasks involving subtle temperature gradient variations; (3) The transfer learning capability enabled by pretrained weights effectively mitigates the limitations caused by the relatively small size of thermal imaging datasets.

During the forward pass, the original thermal image sequence is transformed into a two-dimensional striped grayscale image through thermal image reconstruction and enhancement. The encoder takes the reconstructed thermal signal as input, and the high-level semantic features extracted by the backbone network are subsequently passed to a multilayer perceptron-based regression module. This module includes fully connected layers, nonlinear activation functions (ReLU), and dropout regularization layers, which together enable the final depth regression prediction.

*2.3.2. The thermal signal reconstruction and enhancement block*

This study analyzes pixel-level grayscale response curves during thermal excitation, exhibiting typical thermal diffusion: rapid heating rise followed by gradual cooling decline. Pixel responses vary distinctly with defect depth—heating slopes, peak values, and cooling decay rates all correlate with depth variations. These features provide physical foundations for image-based depth modeling and prediction.

A response analysis was performed to investigate the relationship between frame number and grayscale values for selected pixels. Pixel values were extracted at intervals of 10 frames. After applying denoising and smoothing procedures, Fig.4 illustrates the value evolution curves of nine representative pixels, which correspond to regions with different defect depths throughout the data acquisition process.

The thermogram sequence is mapped into one 2D image. Taking a sample with a defect depth of 0.24 mm as an example, Fig.5(a) shows the thermogram sequence $G_i(n)$ of a specific pixel located at the region with a defect depth of 0.24 mm, where $i = 1, 2, \ldots, 9$ denotes the nine different defect depths, and $1 \leq n \leq 1024$, where $n$ represents the frame index in the sequence. The reconstructed 2D image is shown in Fig.5(b), with its grayscale values given by:

$$G'_i(c,n) = G_i(n) \quad (1)$$

where $c$ represents the column index of the selected pixel in the image, and $1 \leq c \leq 1024$.

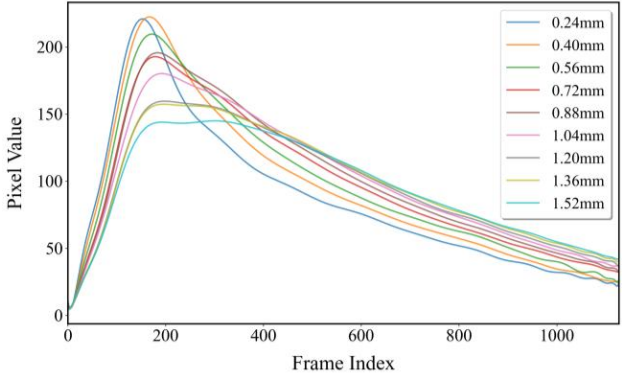

**Fig. 4** Pixel value curves after polynomial fitting.

To further enhance the representation of temperature distribution features, a logarithmic transformation was applied to the image. This transformation compresses the dynamic range of high-temperature regions while amplifying detail in low-temperature areas, thereby making subtle temperature differences—otherwise indistinguishable under a linear scale—more prominent. The result significantly improves the visual expression of thermal anomalies. The involved logarithmic transformation and normalization are defined as follows:

$$q = \ln(v) \quad (2)$$

$$s = 255[q - \min(q)]/[\max(q) - \min(q)] \quad (3)$$

Where $\ln(\,)$ is the natural logarithm function, $v$ is the input pixel value (ranging from 0 to 255), and $s$ is the final output pixel value after transformation and normalization. Fig.5(c-d) present the pseudo-color images corresponding to the 0.24 mm defect depth and its log-enhanced version, respectively.

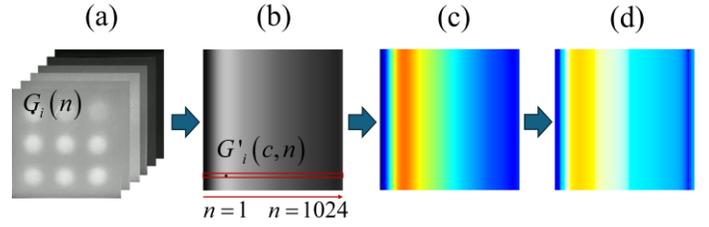

**Fig. 5** (a) A thermogram sequence for the sample with defect depth of 0.24 mm; (b) the reconstructed 2D image; (c) the corresponding pseudo-color image of (b); (d) the log transformed image of (b).

## 2.4. The training settings

The PQM is trained in an end-to-end manner. The optimizer is AdamW, with an initial learning rate set to $1 \times 10^{-3}$ and a weight decay parameter of $10^{-4}$. To enhance convergence, the ReduceLROnPlateau learning rate scheduler is employed. When the validation loss does not decrease for five consecutive epochs, the learning rate is automatically reduced to 50% of its previous value, allowing the model to perform finer optimization during plateau phases. The ReduceLROnPlateau strategy is defined as:

$$\eta_t = \gamma \cdot \eta_{t-1}, \text{ if } L_{val}^{(t)} \leq \min_{k<t} L_{val}^{(k)} + \varepsilon \text{ for } p \text{ epochs} \quad (4)$$

where $\eta_t$ denotes the learning rate at epoch $t$; $\gamma$ represents the scaling factor, $\varepsilon$ the minimum improvement threshold, and $L_{val}^t$ $L_{val}(t)$ is the validation loss at epoch $t$;.

The loss function is defined as a weighted hybrid regression loss function:

$$L_{total} = \lambda \cdot \frac{1}{n}\sum_{i=1}^{n}(\hat{y}_i - y_i)^2 + (1-\lambda) \cdot \frac{1}{n}\sum_{i=1}^{n}|\hat{y}_i - y_i| \quad (5)$$

where $\hat{y}_i$ denotes the predicted defect depth by the model, $y_i$ the ground truth depth label, $n$ the total number of samples, $\lambda$ is the weighting factor.

The dataset constructed in this study comprises 1,773 pixel-level samples from nine defect depth regions, with 197 pixels collected for each region. Compared with conventional datasets, the dataset offers significant advantages: (1) Pixel-based thermal image reconstruction enables the generation of a large amount of training data; (2) The dataset contains rich temporal information, which facilitates the model's ability to learn the evolution patterns of defects. During data preprocessing, all input images are resized to 512×512 pixels and normalized with a mean of 0.5 and a standard deviation of 0.5. The dataset is divided into 70% for training, 15% for validation, and 15% for testing. During training, the batch size is set to 8, and the total number of training epochs is 100. To prevent gradient explosion and ensure training stability, a gradient clipping strategy with a maximum norm of 1.0 is applied.

## 3. EXPERIMENTAL RESULTS

### 3.1. Depth estimation performance

Table 2 summarizes the key regression evaluation metrics of different DNN models on the test set. The compared models include ConvNeXt [31], ViT [31], EfficientNetV2 [32], RegNet [32], ResNet [33], and VAE [34]. The results demonstrate that the proposed PQTNet achieves superior performance across all regression metrics, significantly outperforming the baseline models.

**Table 2** Comparison of the Proposed PQTNet and Six Commonly Used DNN Models

| Method | RMSE (×10$^{-2}$) | MAE (μm) | MAPE (%) | R$^2$ |
|---|---|---|---|---|
| ConvNeXt [31] | 4.78 | 30.3 | 3.52 | 0.987 |
| VIT [35] | 3.79 | 23.6 | 2.73 | 0.992 |
| EfficientNetV2 [32] | <u>2.69</u> | <u>18.4</u> | <u>1.97</u> | <u>0.996</u> |
| RegNet [36] | 3.86 | 27.7 | 3.00 | 0.991 |
| ResNet [33] | 5.04 | 35.4 | 4.09 | 0.985 |
| VAE [34] | 5.54 | 37.6 | 4.25 | 0.982 |
| PQTNet (Ours) | **2.08** | **9.40** | **1.17** | **0.997** |

**Bold** and <u>underlined</u> values are the best and 2$^{nd}$ best in each column.

Tables 3-5 list the evaluation metrics of the compared DNN models at different defect depths, including MAE in micrometer (μm), MAPE in percentage (%), and mean predicted defect depth values in millimeters (mm). The results show that in shallow regions (e.g., 0.24 mm, 0.56 mm, 0.72 mm), although the PQTNet does not achieve the best performance in terms of MAE and MAPE, it still ranks among the top. This is mainly because shallow defects are less affected by lateral heat conduction, resulting in lower noise levels in the input images, which makes it difficult for the proposed PQTNet to show a significant advantage over others. In contrast, in deeper defect regions, the PQTNet significantly outperforms the others in both MAE and MAPE, with a more concentrated prediction distribution. This advantage is mainly attributed to: (1) the more complex temperature gradient variations in deep defects, where the model demonstrates stronger nonlinear mapping capabilities; (2) the improved prediction stability and robustness in deeper regions, effectively reducing prediction uncertainty.

**Table 3** The MAE values (in μm) of different models on the test set with nine defect depths (in mm).

| Method | True defect depth (mm) | | | | | | | | |
|---|---|---|---|---|---|---|---|---|---|
| | 0.24 | 0.40 | 0.56 | 0.72 | 0.88 | 1.04 | 1.20 | 1.36 | 1.52 |
| ConvNeXt | **3.45** | 26.9 | 15.1 | 25.5 | 29.0 | 32.7 | 44.9 | 56.6 | 33.4 |
| VIT | 4.99 | 17.5 | 12.1 | <u>15.6</u> | 18.4 | 18.2 | 51.4 | 56.9 | <u>14.6</u> |
| EfficientNetV2 | 6.57 | **6.98** | **0.83** | 19.0 | 17.9 | <u>15.4</u> | 30.7 | <u>31.2</u> | 28.6 |
| RegNet | 9.40 | 9.07 | 17.7 | 17.3 | <u>14.2</u> | 31.5 | 53.8 | 32.5 | 53.2 |
| ResNet | 13.7 | 19.9 | 14.3 | 31.1 | 27.7 | 31.3 | 55.6 | 33.6 | 89.7 |
| VAE | 19.6 | 19.8 | 12.1 | 16.3 | 15.1 | 37.8 | <u>28.6</u> | 81.7 | 109 |
| Ours | <u>6.51</u> | <u>7.16</u> | <u>2.63</u> | **9.21** | **4.49** | **3.29** | **22.7** | **15.9** | **9.37** |

**Bold** and <u>underlined</u> values are the best and 2$^{nd}$ best in each column.

**Table 4** The MAPE values (in %) of different models on the test set with nine defect depths (in mm).

| Method | True defect depth (mm) | | | | | | | | |
|---|---|---|---|---|---|---|---|---|---|
| | 0.24 | 0.40 | 0.56 | 0.72 | 0.88 | 1.04 | 1.20 | 1.36 | 1.52 |
| ConvNeXt | **1.44** | 6.74 | 2.71 | 3.54 | 3.30 | 3.14 | 3.74 | 4.16 | 2.20 |
| VIT | <u>2.08</u> | 4.38 | 2.16 | <u>2.17</u> | 2.09 | 1.75 | 4.29 | 4.19 | <u>0.96</u> |
| EfficientNetV2 | 2.74 | **1.74** | **0.14** | 2.63 | 2.03 | 1.48 | 2.56 | <u>2.29</u> | 1.88 |
| RegNet | 3.91 | 2.26 | 3.17 | 2.41 | <u>1.61</u> | 3.03 | 4.48 | 2.39 | 3.50 |
| ResNet | 5.73 | 4.98 | 2.56 | 4.32 | 3.14 | <u>3.01</u> | 4.63 | 2.47 | 5.90 |
| VAE | 8.19 | 4.96 | 2.16 | 2.27 | 1.71 | 3.63 | <u>2.38</u> | 6.01 | 7.15 |
| Ours | 2.71 | <u>1.79</u> | <u>0.47</u> | **1.27** | **0.51** | **0.31** | **1.89** | **1.17** | **0.61** |

**Bold** and <u>underlined</u> values are the best and 2$^{nd}$ best in each column.

**Table 5** The average values (in mm) of different models on the test set with nine defect depths (in mm).

| Method | True defect depth (mm) | | | | | | | | |
|---|---|---|---|---|---|---|---|---|---|
| | 0.24 | 0.40 | 0.56 | 0.72 | 0.88 | 1.04 | 1.20 | 1.36 | 1.52 |
| ConvNeXt | <u>0.237</u> | 0.421 | <u>0.558</u> | 0.731 | <u>0.886</u> | <u>1.03</u> | 1.24 | 1.33 | 1.50 |
| VIT | 0.244 | 0.417 | 0.568 | 0.733 | 0.894 | <u>1.03</u> | 1.24 | 1.38 | 1.53 |
| EfficientNetV2 | 0.246 | **0.406** | **0.561** | 0.739 | 0.892 | <u>1.03</u> | **1.22** | **1.36** | <u>1.51</u> |
| RegNet | <u>0.237</u> | <u>0.393</u> | 0.543 | 0.733 | 0.871 | 1.01 | 1.25 | <u>1.35</u> | **1.52** |
| ResNet | **0.239** | 0.413 | 0.556 | 0.751 | 0.859 | 1.02 | 1.26 | 1.34 | 1.43 |
| VAE | 0.259 | 0.419 | 0.563 | **0.721** | 0.873 | 1.01 | **1.19** | 1.28 | 1.41 |
| Ours | 0.246 | <u>0.407</u> | 0.562 | <u>0.714</u> | 0.878 | 1.04 | <u>1.22</u> | <u>1.35</u> | <u>1.53</u> |

**Bold** and <u>underlined</u> values are the best and 2$^{nd}$ best in each column.

## 3.2. Ablation experiments

To further investigate the effectiveness of each component within the PQTNet, multiple Ablation Experiments (AEs) were conducted focusing on the impact of the data augmentation strategy and the RRH on model performance. As summarized in Table 6, the full PQTNet model outperforms the original EfficientNetV2 backbone across all regression metrics, demonstrating that the introduced modules contribute positively to overall prediction accuracy.

**Table 6** Ablation experiments on the PQTNet.

| AE | Data Enhancement | RRH | RMSE (×10⁻²) | MAE (μ m) | MAPE (%) | $R^2$ |
|---|---|---|---|---|---|---|
| ① | × | × | 2.69 | 18.5 | 1.97 | 0.9956 |
| ② | √ | × | 2.53 | 13.8 | 1.40 | 0.9961 |
| ③ | × | √ | 2.62 | 17.1 | 1.69 | 0.9958 |
| ④ | √ | √ | **2.08** | **9.40** | **1.17** | **0.9974** |

Fig. 6 illustrates the comparison of loss curves during training when using the original images and the augmented images as inputs, respectively. Both cases exhibit rapid convergence within the first five epochs. Notably, the model trained with augmented data consistently maintains lower training and validation losses, indicating not only an improved fitting ability but also enhanced generalization performance. Further examination of the validation loss curves reveals that the augmented model exhibits significantly smaller fluctuations compared to the baseline, demonstrating greater training stability. This suggests that the introduction of the logarithmic transformation-based data augmentation strategy serves as an effective regularization method, mitigating the risk of overfitting. Overall, the augmentation strategy expands the diversity of the thermal image data distribution, enabling the model to learn more discriminative and robust feature representations, thereby achieving superior performance on both the training and validation sets.

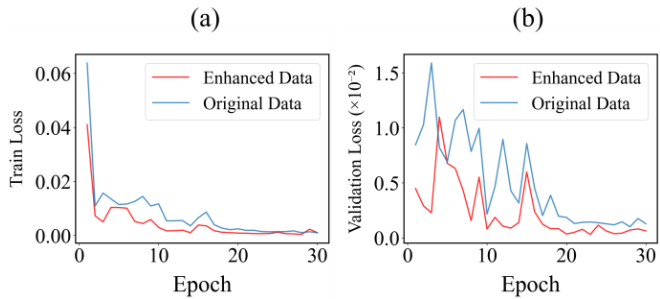

**Fig. 6** Training and validation loss curves of the PQTNet.

## 4. CONCLUSIONS

This paper proposes a novel thermal image sequence processing method that reconstructs raw infrared thermal data along the temporal axis into 2D images, upon which a data augmentation strategy is introduced as model input. Based on EfficientNetV2 as the backbone, a Residual Regression Head (RRH) is designed to construct a DNN model named PQTNet. The regression head employs a three-stage fully connected architecture, integrating learnable scale and shift parameters to perform affine transformation on the regression output, thereby enhancing the model's adaptability to the numerical range of defect depths and improving prediction stability. The PQTNet was evaluated on an in-house active thermography dataset for defect depth estimation and compared against mainstream DNNs. Results demonstrate that it outperforms these models, achieving a depth estimation error of 0.0094 mm with a coefficient of determination $R^2$ exceeding 99%. Future work will proceed along two main directions: firstly, expanding the study to include a broader range of additive manufacturing materials such as CFRP, ABS, and PA, to validate the generalization capability and robustness of the proposed model; secondly, conducting in-depth investigations into the internal heat transfer mechanisms of materials, incorporating heat conduction theory to analyze more complex internal defect geometries, thereby further enhancing the accuracy and reliability of defect depth estimation.